\documentclass[10pt,english]{article}
\usepackage{subfigure}
\usepackage{graphicx}
\usepackage{float}
\usepackage[T1]{fontenc}
\usepackage[latin9]{inputenc}
\usepackage[margin=1.25in]{geometry}
\usepackage{float}
\usepackage{amsmath}
\usepackage{amsbsy}
\usepackage{setspace}
\usepackage{amssymb}
\usepackage{esint}
\usepackage{cite}
\newfloat{algorithm}{tbp}{loa}
\floatname{algorithm}{Algorithm}
\usepackage{hyperref}
\usepackage{listings}

\newlength\myindent
\makeatletter

\floatstyle{ruled}
\newfloat{algorithm}{tbp}{loa}
\floatname{algorithm}{Algorithm}

\begin{document}
%\begin{singlespace}

\title{Residual Random Neural Networks}

\author{M. Andrecut}

\date{}
\date{October 24, 2024}

\maketitle
{

\centering Unlimited Analytics Inc.

\centering Calgary, Alberta, Canada

\centering mircea.andrecut@gmail.com

} 

\begin{abstract}
The single-layer feedforward neural network with random weights is a recurring motif in the neural networks literature. 
The advantage of these networks is their simplified training, which reduces to solving a ridge-regression problem. 
A general assumption is that these networks require a large number of hidden neurons relative to the dimensionality of the data samples, in order to achieve good classification accuracy. 
Contrary to this assumption, here we show that one can obtain good classification results even if the number of hidden neurons has the same order of magnitude as the dimensionality of the data samples, if this dimensionality is reasonably high. 
Inspired by this result, we also develop an efficient iterative residual training method for such random neural networks, and we extend the algorithm to the least-squares kernel version of the neural network model. 
Moreover, we also describe an encryption (obfuscation) method  
which can be used to protect both the data and the resulted network model. 
\end{abstract}

\section{Introduction}

Randomization of the weights in the single-layer feedforward neural network (SLFNN) is a recurring motif in the neural networks research. 
As a consequence, the "benefits of randomization" have been discovered (and rediscovered) many times by different authors in various contexts \cite{key-1}-\cite{key-9}. 
The main idea is the fixed random initialization of the weights in the hidden layer of the SLFNN, such that the training is reduced to computing the 
ridge-regression solution of the linear system corresponding to the hidden layer outputs and the classification targets. 
Therefore, the key advantage of these random neural networks is their simplified training algorithm. Also, it has been shown that despite of the randomness of their hidden weights layer, 
these simple networks are still capable of a good approximation of any non-constant piecewise continuous function \cite{key-5}. 

A general assumption is that these random-SLFNNs require a large number of hidden neurons $J$ relative to the dimensionality of the samples $M$, in order to achieve good classification accuracy. 
Typically, for higher accuracy applications the number of hidden neurons is required to be an order of magnitude larger (or even more) than the dimensionality of the data samples, $J\gg M$ \cite{key-1}-\cite{key-9}. 
Thus, the classification accuracy of random-SLFNNs depends on two parameters, the dimensionality of the space $M$ and the number of the hidden random neurons $J$. 
Therefore, the application of random-SLFNNs to very-high-dimensional data becomes problematic, since by increasing the dimensionality of the data $M$, the 
number of hidden neurons $J$ is also expected to increase by an order of magnitude, or even more. 

The above observation rises the following fundamental question: is there an optimal ratio $J/M$, such that the classification performance is maximized? 
An exact theoretical answer to this question is not obvious because high-dimensional spaces have counter-intuitive geometric properties \cite{key-10}, \cite{key-11}. 
A typical counter-intuitive example is the "concentration of distances", which means that the distances between nearest neighbours of randomly drawn points become large, 
and consequently closer to the distances to the farthest neighbours. This means that the norms of randomly drawn points in a high-dimensional space 
"concentrate", leading to the so called "empty space phenomenon". The first impression is that this property cannot be useful (the so called "curse of dimensionality"), however 
it can be shown that it leads to the surprising phenomenon of the "almost orthogonality" of randomly drawn vectors. 
Also, it can be shown that this "almost orthogonality" is improving by increasing the dimensionality $M$ of the space, and therefore intuitively we expect to achieve 
a better separation of the samples to be classified (the so called "blessing of dimensionality"). 

Using the above intuition, here we derive a simple strategy to improve the classification accuracy of random-SLFNNs by increasing the dimensionality of the data samples $M$.  
This approach results in a reduced number of required hidden neurons $J$, such that both $M$ and $J$ have the same order of magnitude. 
We further exploit this idea by developing an efficient and low complexity iterative method, that can be used to train a 
residual random neural network (RRNN) model, significantly improving their classification accuracy. 
Following these results, we also extend the residual training algorithm to the least-squares kernel version of the neural network (RRKN), achieving a similar classification accuracy, with a more simplified architecture. 
Moreover, we also describe an encryption (obfuscation) method based on orthonormal random matrices, 
which can be used to protect both the data and the resulted network model. 

\section{Random-SLFNN}

Let us assume that $X= [x_{n,m}]_{N\times M}$ is the training data matrix, such that each row vector is a data sample $x_{n}=(x_{n,0},x_{n,1},...x_{n,M-1}) \in \mathbb{R}^M$ from one of the $K$ classes. 
The classification problem requires the mapping of the row vectors of a new, unclassified data matrix $\tilde{X}=[\tilde{x}_{nm}]_{N'\times M}$, to the corresponding classes $\{0,1,...,K-1\}$. 

In the random-SLFNN approach, the $X$ and $\tilde{X}$ matrices are encoded using a random projection matrix $U= [u_{j,m}]_{J\times M}$, typically drawn from the normal distribution $u_{j,m}\in \mathcal{N}(0,1)$:
\begin{equation}
H = h(XU^T), \quad \tilde{H} = h(\tilde{X}U^T), 
\end{equation}
where $h$ is a non-linear activation function, and $J$ is the random encoding dimension, which (as mentioned before) it is typically assumed to be large, $J\gg M$. 

The weights of the output layer are determined by solving the ridge regression problem  \cite{key-1}-\cite{key-9}:
\begin{equation}
W = \text{arg}\min_W \left\lbrace  \Vert HW - Y \Vert^2 + \lambda \Vert W \Vert^2 \right\rbrace ,
\end{equation}
such that, the weights $W$ are:
\begin{equation}
W = (HH^T + \lambda I)^{-1}H^TY.
\end{equation}
Here, $I$ is the $J \times J$ identity matrix, and $\lambda\geq 0$ is the regression parameter.
Also, each row $y_{n} \in \mathbb{R}^K$ of $Y$ corresponds to the class of the data point $x_n$. 

The classes $C_k$ are encoded into the matrix $Y$ using the one-hot encoding approach: 
\begin{equation}
x_n \in C_k \Leftrightarrow y_{n,i} = 
  \begin{cases}
   1  & \text{if } i = k \\
   0 & \text{otherwise }
  \end{cases}.
\end{equation}
Therefore, in order to classify the rows of the test data matrix $\tilde{X}$ we can use the following simple criterion: 
\begin{equation}
\tilde{x}_i \in C_k \quad \text{if} \quad k = \text{arg}\max_{k=0.1,...,K-1} \tilde{y}_{i,k}, \quad i=0,1,...,N'-1,
\end{equation}
where
\begin{equation}
\tilde{Y} = \tilde{H}W= [\tilde{y}_{i,k}]_{N' \times K}.
\end{equation}

\section{Random Projections Geometry}

\subsection{Volume concentration}

Since the data samples to be classified are points in the $M$ dimensional space, here we will discuss in more detail some of the geometric properties of $\mathbb{R}^M$ related to random projections \cite{key-12}. 

The hypersphere is a generalization of the sphere to $\mathbb{R}^M$, and its points have the same property of being 
situated at the same distance $R>0$ from its center $(0,...,0)$:
\begin{equation}
\mathcal{S}^{M-1}(R) = \{x \in \mathbb{R}^M \mid \Vert x \Vert = R \}.
\end{equation}
The hypersphere and its interior points correspond to an $M-$dimensional object called ball: 
\begin{equation}
\mathcal{B}^{M}(R) = \{x \in \mathbb{R}^M \mid \Vert x \Vert \leq R \}.
\end{equation}
The unit hypersphere and the unit ball have a radius $R=1$. Any ball of radius $R>0$ can be mapped into a unit ball using the following simple scaling transformation $x \leftarrow x/R$.

Let us denote the volume and the area of the ball $\mathcal{B}^{M}(R)$ by $V_M(R)$, and respectively $A_{M-1}(R)$. 
As a consequence of the scaling transformation mentioned above, the ball $\mathcal{B}^{M}(R)$ is obtainded from the unit ball $\mathcal{B}^{M}(1)$ by scaling with $R$ all the $M$ directions $x_m$, 
and therefore $V_M(R)$ must be proportional to $R^M$. Also, the volume can be calculated by integrating the infinitesimal volume element $dV = dx_1...dx_M$ over the region in 
$\mathbb{R}^n$ constrained by $\Vert x \Vert \leq R$: 
\begin{equation}
V_M(R) = \int ... \int_{x_1^2+...+x_M^2 \leq R^2} dx_1...dx_M = V_M(1)R^M,
\end{equation}
where $V_M(1)$ is the the volume of the unit ball. 

Alternatively, the volume of the ball can be calculated by integrating the infinitesimal shell volume $dV(r) = A_{M-1}(r)dr$: 
\begin{equation}
V_M(R) = \int_0^R A_{M-1}(r)dr.
\end{equation}
Reciprocally we have:
\begin{equation}
A_{M-1}(r) = \frac{dV_{M}(r)}{dr} = MV_M(1)r^{M-1}=A_{M-1}(1)r^{M-1},
\end{equation}
where $A_{M-1}(1)$ is the area of the unit ball.

Let us now consider the following Gaussian function:
\begin{equation}
f(x_1,...,x_M) = e^{-(x_1^2+...+x_M^2)}.
\end{equation}
The integral of $f$ over the whole space $\mathbb{R}^n$ can be easily calculated as a product of Gaussian integrals of single variables:
\begin{equation}
I_M = \int_{\mathbb{R}^M} fdV = \prod_{m=1}^M \int_{-\infty}^{\infty}e^{-x_m^2}dx_m = \pi^\frac{M}{2}.
\end{equation}
The same integral $I_M$ can be also calculated using the polar coordinates:
\begin{align}
r^2 &= x_1^2+...+x_M^2,\\
dV(r) &= A_{M-1}(r)dr = A_{M-1}(1)r^{M-1}dr,
\end{align}
as following:
\begin{equation}
I_M = A_{M-1}(1) \int_0^\infty r^{M-1} e^{-r^2}dr = \frac{1}{2} A_{M-1}(1) \Gamma\left( \frac{M}{2} \right),
\end{equation}
where $\Gamma$ is the Gamma function.

From here we can find the area and the volume of the unit ball:
\begin{equation}
A_{M-1}(1) = \frac{2\pi^{\frac{M}{2}}}{\Gamma(\frac{M}{2})}, \quad V_M(1) = \frac{2\pi^{\frac{M}{2}}}{M\Gamma(\frac{M}{2})}.
\end{equation}

One can see that both the area and the volume of the unit ball tend to zero as the dimension $M$ of the space increases:
\begin{equation}
\lim_{M\rightarrow\infty} A_{M-1}(1) = 0, \quad \lim_{M\rightarrow\infty} V_{M}(1) = 0,
\end{equation}
because 
\begin{equation}
\Gamma \left(  \frac{M}{2} \right)  = \sqrt{\pi}\frac{(M-2)!!}{2^{\frac{M-1}{2}}},
\end{equation}
and the the exponential grows slower than the double factorial. 
This is a counter intuitive result, which will be explained next. 

Let us consider two concentric balls $\mathcal{B}^M(1)$ and $\mathcal{B}^M(1-\varepsilon)$, $0<\varepsilon \ll 1$. 
The ratio $\gamma_M(1)$ of the volume of the thin spherical shell with the width $\varepsilon$, 
and of the unit ball $\mathcal{B}^M(1)$ tends to one, as $M\rightarrow\infty$ \cite{key-10}, \cite{key-11}, \cite{key-12}:
\begin{equation}
\lim_{M\rightarrow\infty}  \gamma_M(1) = \lim_{M\rightarrow\infty} \frac{V_M(1) - V_M(1)(1-\varepsilon)^M}{V_M(1)} = \lim_{M\rightarrow\infty} [1-(1-\varepsilon)^M] = 1.
\end{equation}
This means that a very thin spherical shell, with a width $0<\varepsilon \ll 1$, near the surface, essentially contains the whole volume of the ball $\mathcal{B}^M(1)$. 
Therefore, the volume of the unit ball is concentrated in a very narrow shell at the surface, and this explains why the volume and the area of the 
unit ball tend to zero in a high dimensional space.

\subsection{Random points on the hypersphere}

As a consequence of the volume concentration result, the points $x\in \mathbb{R}^M$ with the coordinates drawn from the normal distribution $\mathcal{N}(0,1)$ in a high dimensional space,
are concentrated very close to the hypersphere $S^{M-1}(\sqrt{M})$, such that $\Vert x \Vert \simeq \sqrt{M}$.

Let us now consider two random vectors $u,v\in \mathbb{R}^M$, with the coordinates drawn from the normal distribution $\mathcal{N}(0,1)$, such that $\Vert u \Vert \simeq \Vert v \Vert \simeq \sqrt{M}$ . 
Since the rotations are orthogonal transformations they do not change the norms (distances), or the angles. 
Thus, by rotating the coordinate system, the first vector can be positioned such that $u\simeq(1,1,...,1)$.  
Assuming that after the rotation, the second vector has the coordinates $v=(v_1,v_2,...,v_M)$, then the distance between the two points is:
\begin{equation}
d_M(u,v)=\Vert u-v \Vert = \sqrt{\sum_{m=1}^M (v_m -1)^2} = \sqrt{2M}\sqrt{1-\left\langle v \right\rangle} \simeq  \sqrt{2M},
\end{equation}
where $\left\langle v \right\rangle = M^{-1}\sum_{m=1}^M v_m$ is the mean of the coordinates of $v$, and for large $M$ we have  $\left\langle v \right\rangle \rightarrow 0$. 
On the other hand, we also have:
\begin{equation}
d_M^2(u,v) = \Vert u-v \Vert^2 = \Vert u \Vert^2 + \Vert v \Vert^2 - 2 u \cdot v \simeq M + M - 2 u \cdot v \simeq 2M,
\end{equation}
which means that the dot product of the two vectors is: 
\begin{equation}
u \cdot v \simeq 0.
\end{equation}
Thus, any two random vectors $u,v \in \mathbb{R}^M$ with the coordinates sampled from the normal distribution $\mathcal{N}(0,1)$ are almost orthogonal, if the dimensionality $M$ of the space is large. 
In general, one can show that in a high dimensional space there is an exponentially large number of "almost orthogonal" randomly chosen vectors. 
For example, using the Chernoff bound\cite{key-13} one can show that the probability that two such random vectors $u$ and $v$ are "almost orthogonal":
\begin{equation}
\vert u \cdot v \vert < \delta\ll 1,
\end{equation}
is:
\begin{equation}
\text{Pr}_{\delta}(u \perp v) > 1 - \exp \left( -M \delta^2 \right).
\end{equation}

In what follows we rely on this counter intuitive property, and we show numerically that by increasing the dimensionality $M$ of the space we get a better separation of the samples, and consequently a better 
classification accuracy, as pointed out by the previous results \cite{key-10}, \cite{key-11}. 

\section{Preliminary Numerical Results}

\subsection{Data}

In order to illustrate numerically the proposed method, here we use two well known data sets: MNIST \cite{key-14} and fashion-MNIST (fMNIST) \cite{key-15}. 

The MNIST data set is a large database of handwritten digits $\{0,1,...,9\}$, containing 60,000 training images 
and 10,000 testing images. These are monochrome images with an intensity in the interval $[0,255]$, and the size of $M=28 \times 28 = 784$ pixels. 
The fashion-MNIST dataset also consists of 60,000 training images and a test set of 10,000 images. The images are also monochrome, with an intensity in the interval $[0,255]$ and the size of $M=28 \times 28 = 784$ pixels. 
However, the fashion-MNIST is harder to classify, since it is a more complex dataset, containing images from $K=10$ different apparel classes: 
0 - t-shirt/top; 1 - trouser; 2 - pullover; 3 - dress; 4 - coat; 5 - sandal; 6 - shirt; 7 - sneaker; 8 - bag; 9 - ankle boot. 

To further simplify the computations, we normalize both the data and the random projection vectors as following:
\begin{align}
x_n &\leftarrow x_n - \langle x_n \rangle, \quad x_n \leftarrow x_n/\Vert x_n \Vert, \\
\tilde{x}_n &\leftarrow \tilde{x}_n - \langle \tilde{x}_n \rangle, \quad \tilde{x}_n \leftarrow \tilde{x}_n/\Vert \tilde{x}_n \Vert\\
u_j &\leftarrow u_j/\Vert u_j \Vert
\end{align}
such that all the vectors involved in the computation lie on the $M-$dimensional unit hypersphere $\mathcal{S}^{M-1}(1)$ (an optional initial step which may improve the classification is also $x_n \leftarrow \sqrt{x_n}$ \cite{key-16}).
Thus, the random projection of a data sample is simply the cosine of the angle between the sample and the random projection vector. 
Also, the non-linear activation function used in all the numerical experiments is the hyperbolic tangent. 
Therefore, after accounting for the above normalization procedure, the matrices $H$ and $\tilde{H}$ from random-SLFNN equations 
can be rewritten as following:
\begin{equation}
H = \text{tanh}(\sqrt{M}XU^T), \quad \tilde{H} = \text{tanh}(\sqrt{M}\tilde{X}U^T). 
\end{equation}

\subsection{Preliminary results}

In this preliminary experiment we separately increase the number of projections $J$, and the size of the samples $M$. 
In the first case we keep the size of the images constant $M=L^2=\ell^2$, and we increase the number of random projections $J=qM=qL^2=q\ell^2$, $q=1,5,10,15,20$. 
In the second case we increase the size of the images $M=L^2=s^2\ell^2$, $s=1,2,3,4,5$, and we set the number of projections to $J=M=L^2=s^2\ell^2$. 
In order to resize the images we use the "resize" method from the "scikit-image" package in Python.
Also, in this experiment the regularization parameter was set to $\lambda = 0$ for MNIST, and respectively $\lambda = 10$ for fMNIST. 
The classification results are shown in Table 1. We notice that the classification accuracy values are in fact better in the case where 
we increase the size of the images, and we keep the number of random projections equal to the size of the images. 

\begin{table}[ht]
\centering
\caption{Preliminary random-SLFNN accuracy results $(\%)$.}
\begin{tabular}[t]{|l|c|c|l|c|c|}
\hline
\multicolumn{3}{|c|}{$J=qM, L=\ell$} & \multicolumn{3}{|c|}{$J=M, L=s\ell$} \\
\hline
$q$ & MNIST & fMNIST & $s$ & MNIST & fMNIST\\
\hline
1 & 93.61 & 85.43 & 1 & 93.61 & 85.43\\
5 & 97.21 & 88.40 & 2 & 97.57 & 88.28\\
10 & 97.86 & 89.51 & 3 & 98.09 & 89.29\\
\hline
\end{tabular}
\end{table}

The obtained results are surprising, and they further suggest that we can also try to combine the two options, by increasing both the size of the images and the number of random projections. 
In Table 2, we show the results obtained for this case. One can see that if we increase the size of the images we do not need to increase the number the random projections too much 
in order to obtain comparable (or even better) accuracy results, which contradicts the belief that a good accuracy requires a very large number of random projections,  relative to the image size.

\begin{table}[ht]
\centering
\caption{Preliminary random-SLFNN combined accuracy results $(\%)$, $J=qM, L=s\ell$.}
\begin{tabular}[t]{|l|l|c|c|}
\hline
$s$ & $q$ & MNIST & fMNIST \\
\hline
1 & 1 & 93.61 & 85.43 \\
1 & 2 & 95.25 & 86.58 \\
2 & 1 & 97.57 & 88.28 \\
2 & 2 & 98.14 & 89.40 \\
\hline
\end{tabular}
\end{table}

\section{Adding More Information}

\subsection{Additional information}

Following the conclusion of the preliminary numerical experiments, we hypothesize that by adding more information about the data samples, such that we can further increase 
the size of each sample, then we can expect an increase in the classification accuracy. 
A simple way to add more information is to compute some non-linear function of the data sample, and appended it to the initial data sample. 
Obviously, such a function must be non-linear, since a linear function will make the data matrix singular. For example, a simple approach is to add  
the absolute value of the fast Fourier transform (FFT) \cite{key-16}:
\begin{align}
x'_n &\leftarrow \vert \text{FFT}(x'_n - \langle x'_n \rangle ) \vert,\\
x'_n &\leftarrow x'_n/\Vert x'_n \Vert.
\end{align}
Thus, for each data sample we calculate the above absolute FFT value and we append it to the initial data, resulting in a row data vector $(x_n,x'_n)$, with a length of $3M/2$.
We should note that the image resizing is performed before appending the additional information, and the random projections are applied after appending the additional information. 

\subsection{Random-SLFNN with additional information} 

Let us now repeat the previously described experiment, by adding the additional FFT information. The results shown in Table 3 are surprisingly good, reaching 
an accuracy of $\eta=98.63\%$ for MNIST, and respectively $\eta=90.13\%$ for fMNIST, when $s=2$ and $q=2$. 
In all the considered cases the regularization parameter was set to $\lambda=0$ for MNIST, and $\lambda=10$ for fMNIST.

\begin{table}[!h]
\centering
\caption{accuracy $(\%)$ for random-SLFNN with additional FFT information, for $J=qM$, $L=s\ell$.}
\begin{tabular}[t]{|l|l|c|c|}
\hline
$s$ & $q$ & MNIST & fMNIST\\
\hline
1 & 1 & 96.03 & 86.42\\
1 & 2 & 97.08 & 87.91\\
2 & 1 & 98.08 & 89.16\\
2 & 2 & 98.63 & 90.13\\
\hline
\end{tabular}
\end{table}

\section{RRNN Approach}

\subsection{The algorithm}

The main mechanism of the random-SLFNN method is practically a ridge regression of the data samples $H$, fitting the target matrix $Y$. Therefore, once we 
compute the predicted values $\tilde{Y}$, we can subtract these values from the target matrix $Y$, and we can try to apply the method again 
but this time on the residual $R= Y - \tilde{Y}$. Then we can repeat the process until some stopping criterion is satisfied, such as 
$R=\Vert Y - \tilde{Y} \Vert < \varepsilon$, or if the number of iterations exceeds a given limit. The algorithm can be formulated as following: 

\begin{enumerate}
\item Set the maximum number of iterations $T$, the residual limit $\varepsilon > 0$, the parameter $0 <\mu \leq 1$, and the seed of the random number generator.
\item Set the iteration index to $t=0$, and the initial residual $R_0=Y$.
\item Generate a random projection matrix: $U_t$.
\item Compute the data projection matrix: $H_t=h(XU^T_t)$.
\item Solve the regression problem: $W_t = (H_tH_t^T + \lambda I)^{-1}H_t^TR_t$.
\item Compute the residual: $R_{t+1} = R_t - \mu H_tW_t$
\item If $\Vert R_{t+1} \Vert < \varepsilon$ or $t+1 > T$ then set $T^*\leftarrow t$ and stop, \\ otherwise set $t\leftarrow t+1$ and go to 3.
\item Return $U_t$ and $W_t$ for $t=0,1,...,T^*$. 
\end{enumerate}

One can see that at each time step $t=0,1,...,T^*$ we compute a new random projection matrix, and a new residual. 
The final approximation $\hat{Y}$ can be used to find the target approximation: 
\begin{equation}
\tilde{Y} = \mu \sum_{t=0}^{T^*} \tilde{F}_t W_t, 
\end{equation}
where $\tilde{F}_t = h(\tilde{X}U_t^T)$, and $\tilde{X}$ is the test data matrix. 

Since the projection matrices $U_t$ are randomly generated, we do not have to store them, we only need the random seed and they can be computed on the fly. 
However, we do need to store the weights $W_t$. These matrices are not large, and therefore they do not require too much memory for storage. The iterative method can be applied 
also when additional information is added to the data, since in this case we just have to append the additional information to the data matrices $X$, and respectively $\tilde{X}$. 

We should note that by using the iterative algorithm we practically train an error correcting neural network, which we called Residual Random Neural Network (RNN). 
If $T^*$ is the number of steps at the end of the iteration, then the training process will produce 
a RRNN with $T^*$ hidden layers.  
However, the RRNN architecture is completely different from the traditional deep neural network (DNN). In a DNN the information flows sequentially, that is the output of a 
layer is the input of the next layer. In the RRNN architecture each layer has the same input data, and the role of the layers is to correct the output of the previous layers, 
by minimizing the residual of the target values.  
Thus, each new layer added to the RRNN is architecturally identical to the previous layers.  
Also, these RRNNs are "cheaper" to train, comparing to the deep neural networks, which are trained with the backpropagation algorithm.  

\subsection{RRNN numerical results}

In Figures 1 and 2 we show the numerical results obtained using the RRNN approach for the MNIST, and respectively the fMNIST datasets. One can see that for $s=2$, $q=2$ and $T^*=15$ iteration steps (or layers), 
the classification accuracy reaches $\eta \simeq 99.12 \%$ for MNIST ($\lambda =0$, $\mu=1/2$), and respectively $\eta \simeq 91.41 \%$ for fMNIST ($\lambda =10$, $\mu=1/2$). 
These results are surprisingly good for such a simple method, which as shown above is actually based on an iterative regression of random projections. 

\begin{figure}[!ht]
\centering \includegraphics[width=7.5cm]{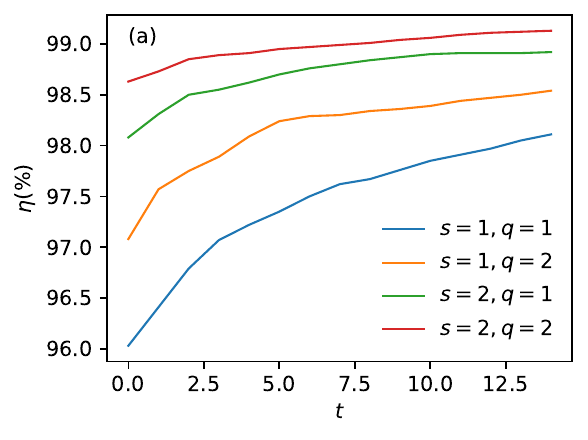}
\centering \includegraphics[width=7.5cm]{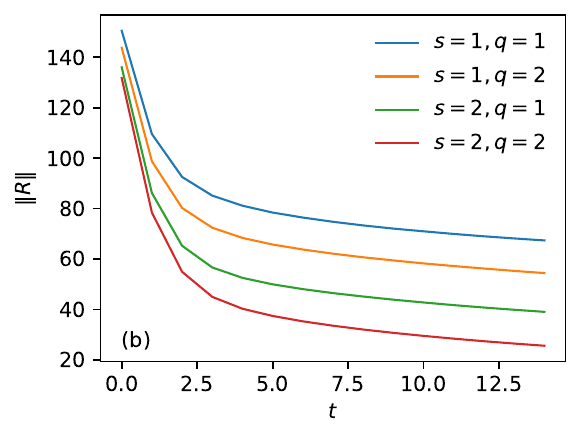}
\caption{RRNN results for MNIST: (a) accuracy; (b) residual norm.}
\end{figure}

\begin{figure}[!ht]
\centering \includegraphics[width=7.5cm]{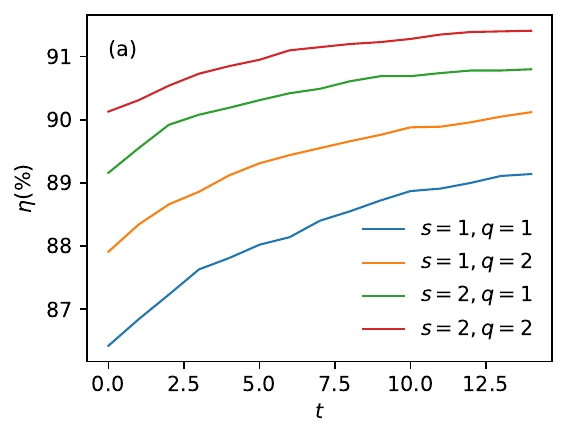}
\centering \includegraphics[width=7.5cm]{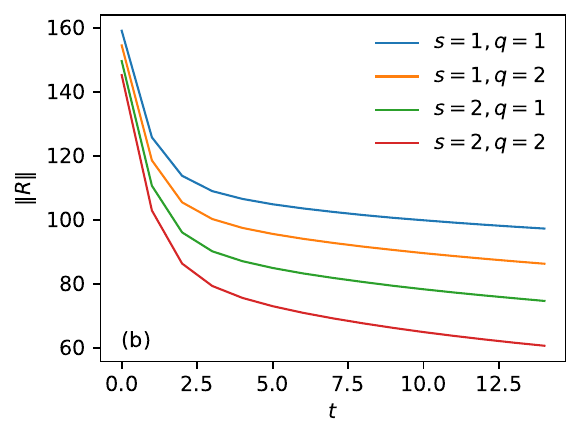}
\caption{RRNN results for fMNIST: (a) accuracy; (b) residual norm.}
\end{figure}

\section{Randomized kernel method}

Kernel methods are used to transform a given non-linear problem into a linear one by using a similarity kernel function $\Phi(x,x')$ defined over pairs of input data points $(x,x')$ \cite{key-17}. 
This way, the input data $x$ is mapped into a feature space $\phi(x)$, 
where the inner product $\langle\cdot,\cdot\rangle$ can be calculated with a 
positive definite kernel function, such that the mapping is done implicitly:
\begin{equation}
\Phi(x,x') = \langle \phi(x),\phi(x')\rangle.
\end{equation}
An important result is that any non-linear function $f$ can be expressed as a linear combination of kernel products evaluated on the 
training data points $\chi = \{x_n|n=1,\dots,N \}$ \cite{key-17}:
\begin{equation}
f(x) = \sum_{n=1}^{N} a_n \Phi(x,x_n).
\end{equation}

A randomized kernel is obtained by randomly selecting a block of $J$ training samples $X_J \subset X$, and then computing the $N\times J$ kernel matrix \cite{key-16}:
\begin{equation}
\Phi(X,X_J) = \varphi(\langle X,X_J \rangle),
\end{equation}
where $\varphi$ is an element wise nonlinear function of the inner product:
\begin{equation}
\langle X,X_J \rangle = XX_J^T.
\end{equation}
The weights of the output layer are determined by solving the least-squares regression problem:
\begin{equation}
W_J = \text{arg}\min_{W_J} \left\lbrace  \Vert \Phi(X,X_J)W_J - Y \Vert^2 + \lambda \Vert W_J \Vert^2 \right\rbrace ,
\end{equation}
such that, the weights $W_J$ are:
\begin{equation}
W_J = \left[\Phi(X,X_J)\Phi(X,X_J)^T + \lambda I \right] ^{-1}\Phi(X,X_J)^TY.
\end{equation}
Here, $I$ is the $J \times J$ identity matrix, $\lambda\geq 0$ is the regression parameter, and 
each row $y_{n} \in \mathbb{R}^K$ of $Y$ corresponds to the class of the data point $x_n$. 

In the case of RRKN the classes are encoded using the same one-hot encoding procedure (4)-(5), however 
in this case the target approximations are given by:
\begin{equation}
\tilde{Y} = \Phi(\tilde{X},X_J)^T W_J= [\tilde{y}_{i,k}]_{N' \times K}.
\end{equation}
We should emphasize that in the classification step, $\Phi(\tilde{X},X_J)$ is the kernel matrix computed using the test data $\tilde{X}$ and the same randomly selected block of training data $X_J$. 

\section{RRKN approach}

\subsection{The algorithm}

As in the case of RRNNs, the main mechanism of the residual approach is to iteratively subtract the predicted values $\hat{Y}$ from the target values $Y$, and then to re-apply the method again on the resulted residual:

\begin{enumerate}
\item Set the maximum number of iterations $T$, the residual limit $\varepsilon > 0$, the number $J$ of the samples in the random kernel, 
the regression parameter $\lambda \geq 0$, and the seed of the random number generator.
\item Set the iteration index to $t=0$, and the initial residual $R(0)=Y$.
\item Randomly select the block matrix $X_J(t)$ from $X$.
\item Compute the kernel matrix: $\Phi(X,X_J(t))$. Here we use a polynomial kernel with $\Phi(X,X_J(t))= \langle X,X_J(t)\rangle^{(k)}$, where $(k)$ means element-wise power. Best results were obtained for $k=5$.
\item Solve the regression problem:
\begin{equation*}
W_J(t) = \left[\Phi(X,X_J(t))\Phi(X,X_J(t))^T + \lambda I \right] ^{-1}\Phi(X,X_J(t))^TR(t)
\end{equation*}
\item Compute the residual: $R(t+1) = R(t) - \Phi(X,X_J(t))W_t$
\item If $\Vert R(t+1) \Vert < \varepsilon$ or $t+1 > T$ then set $T^*\leftarrow t$ and stop, \\ otherwise set $t\leftarrow t+1$ and go to 3.
\item Return $W_J(t)$ for $t=0,1,...,T^*$.
\end{enumerate}

One can see that at each time step $t=0,1,...,T^*$ we compute a new kernel matrix using a randomly selected projection matrix $X_J(t)$, and a new residual $R(t+1)$. 
Finally, the weight matrices $W_J(t)$ can be used to find the target approximation using the $\tilde{X}$ test data matrix: 
\begin{equation}
\tilde{Y} = \sum_{t=0}^{T^*} \Phi(\tilde{X},X_J(t))W_t.
\end{equation}

We notice again that since the kernel projection matrices $X_J(t)$ are randomly selected, we do not have to store them, we only need the random seed. 
However, as in the case of RRNNs we do need to store the weights $W_J(t)$. Thus, if $T^*$ is the number of steps at the end of the iteration, then the training process will produce a RRKN with $T^*$ corrective layers. 

\subsection{RRKN numerical results}

In the first experiment we scale the kernel matrices as following $J=s1000$, $s=1,2,5,10,15$. For each $J$ value, we limit the number of approximation steps (layers) to $T=20$, and 
we let be their index $t \in \{1,2,...,20\}$. 
The results for the MNIST data are shown in Figure 3, and respectively for the fMNIST data in Figure 4.
One can see that the classification accuracy improves gradually by increasing the number of approximation layers $T$, and the size $J$ of the kernel matrix, 
reaching the best values for $T=20$ and $J=15000$: $98.87\%$ for MNIST, and respectively $91.04\%$ for fMNIST. 

\begin{figure}[!ht]
\centering \includegraphics[width=7.5cm]{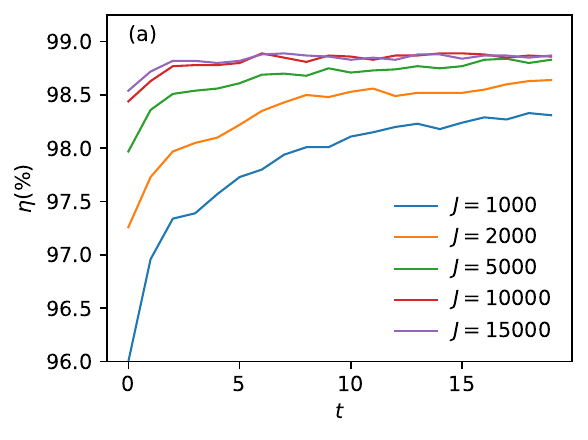}
\centering \includegraphics[width=7.5cm]{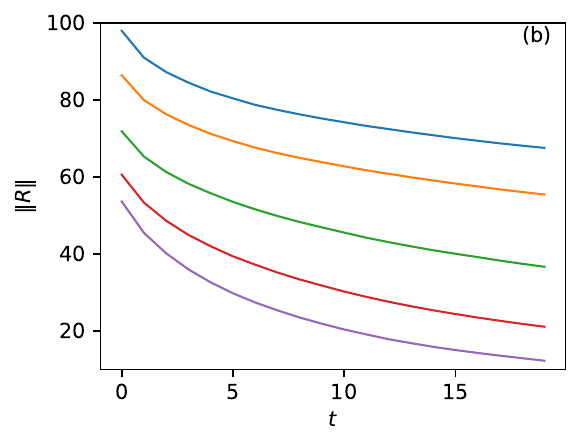}
\caption{Initial RRKN results for MNIST: (a) accuracy; (b) residual norm.}
\end{figure}

\begin{figure}[!ht]
\centering \includegraphics[width=7.5cm]{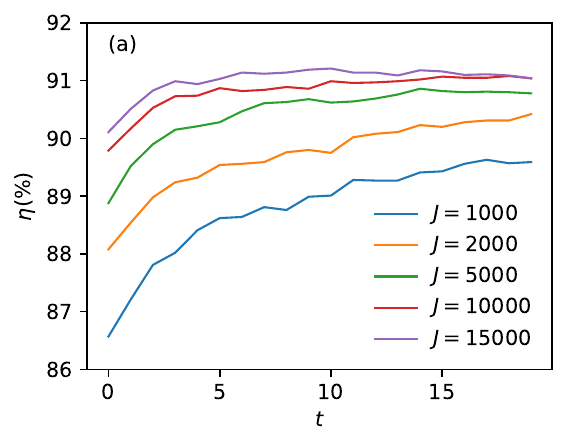}
\centering \includegraphics[width=7.5cm]{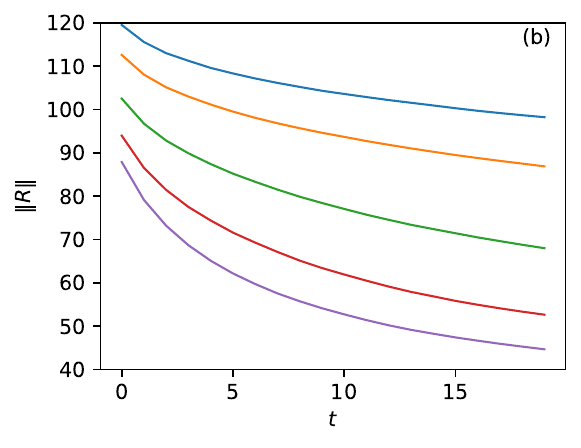}
\caption{Initial RRKN results for fMNIST: (a) accuracy; (b) residual norm.}
\end{figure}

In the second experiment we increase the size of the images by a factor of two, using the same "resize" method from the "scikit-image" package in Python. 
The results for the MNIST data are shown in Figure 5, and respectively for the fMNIST data in Figure 6.
The classification accuracy again improves, and in this case the best results are $99.04\%$ for MNIST, and respectively $91.34\%$ for fMNIST.

\begin{figure}[!ht]
\centering \includegraphics[width=7.5cm]{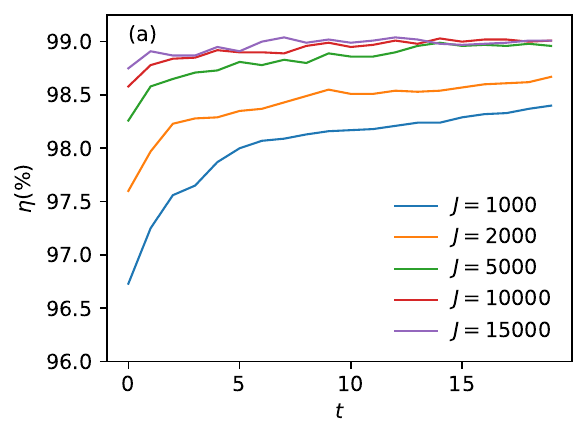}
\centering \includegraphics[width=7.5cm]{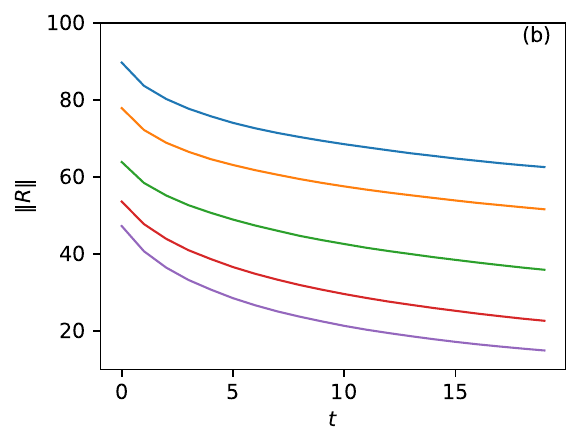}
\caption{Image resizing RRKN results for MNIST: (a) accuracy; (b) residual norm.}
\end{figure}

\begin{figure}[!ht]
\centering \includegraphics[width=7.5cm]{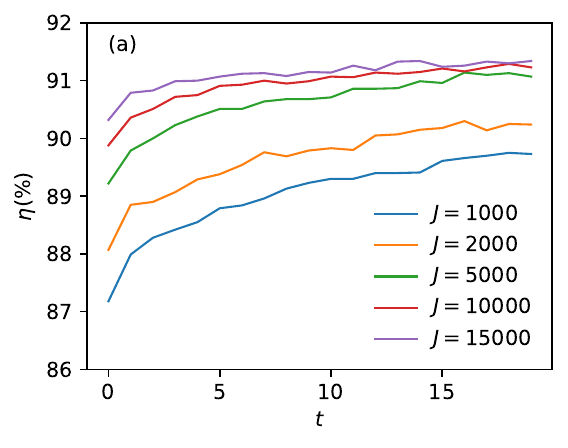}
\centering \includegraphics[width=7.5cm]{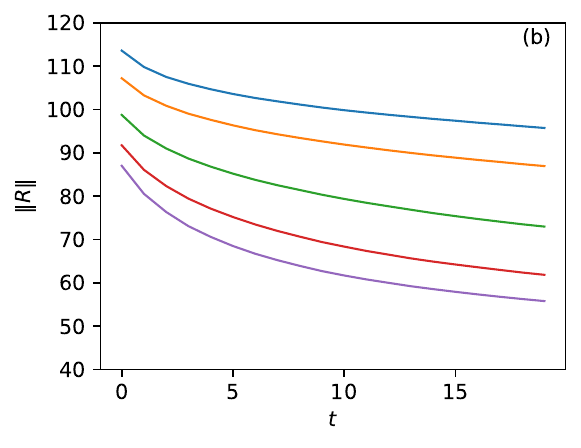}
\caption{Image resizing RRKN results for fMNIST: (a) accuracy; (b) residual norm.}
\end{figure}

As mentioned in the case of RRNNs, instead of simply resizing the images, a better approach is to append more information to them.  
Therefore, we add again the absolute value of the fast Fourier transform (FFT) to each image (30)-(31), resulting in a row data vector $(x_n,x'_n)$, with a length of $3M/2$.
We should emphasize that in this case the length of the data vectors is $3M/2$, which is much smaller than length resulted in the second experiment (Figures 5 and 6), 
where we basically quadrupled the length of the images, to $4M$, as a consequence of the scaling procedure. 
Obviously, adding new information to each data sample (in order to increase dimensionality) has a more significant effect than simply resizing the images. 

In this case, the numerical results for the same parameters are shown in Figure 7 for MNIST, and respectively in Figure 8 for fMNIST. 
One can see that the best results are $99.11\%$ for MNIST and $91.43\%$ for fMNIST, which are comparable to the results previously reported for the 
RRNNs. Also, one can see that in all cases, the classification accuracy of the residual method is starting to saturate for $J> 10000$ and $t> 15$.

\begin{figure}[!ht]
\centering \includegraphics[width=7.5cm]{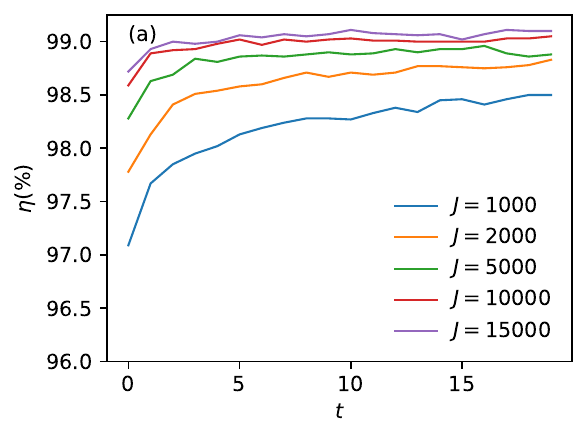}
\centering \includegraphics[width=7.5cm]{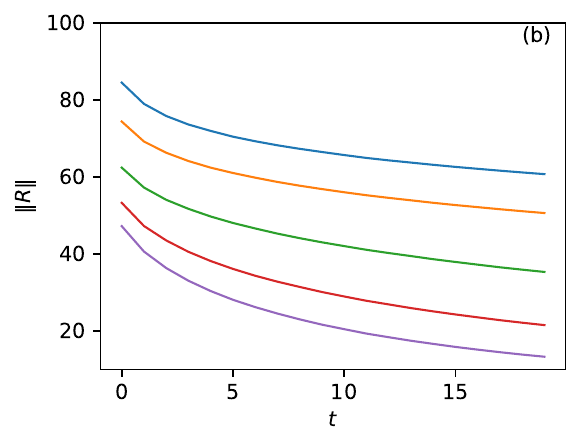}
\caption{RRKN results for MNIST with additional FFT data: (a) accuracy; (b) residual norm.}
\end{figure}

\begin{figure}[!ht]
\centering \includegraphics[width=7.5cm]{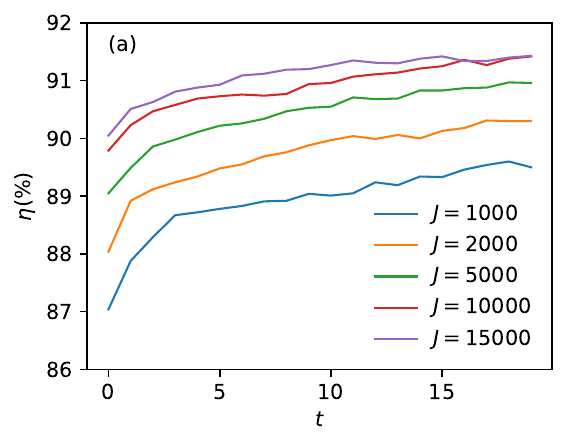}
\centering \includegraphics[width=7.5cm]{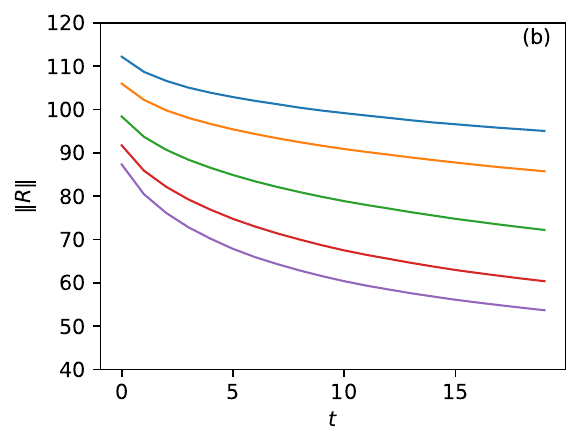}
\caption{RRKN results for fMNIST with additional FFT data: (a) accuracy; (b) residual norm.}
\end{figure}

\section{Security approach}

An interesting aspect of randomness is that one can also use it to derive a simple secure scheme to train and run the RRNN/RRKN model in the cloud, and later to provide access to the model. 
More exactly, we can encrypt (obfuscate) the data before we train the models.  
The process consists in simply obfuscating the data using a secret random orthonormal matrix $Q$. Since the data is $M$ dimensional, the secret matrix $Q$ should be   
a random orthonormal $M\times M$ matrix. For example, such a matrix can be generated using the QR decomposition of a randomly generated matrix $\Psi$ with the elements drawn from the normal distribution \cite{key-18}, 
such that: $\Psi = QR$. 
The result of the decomposition is an orthonormal random matrix $Q$, and a triangular matrix $R$ which can be discarded. 
Since the data $X$ lives in a high dimensional space $\mathbb{R}^M$, we can encrypt (obfuscate) it before we train the RRNN/RRKN model (or any other neural network), as following: $E(X) = XQ$. 
In order to be able to use the model, the test data vector $\tilde{x}$ submitted for classification should also be encrypted using the same secret matrix: $Q\tilde{x}=\tilde{x}'$ (here $\tilde{x}$ and $\tilde{x}'$ are column vectors). 
This works because orthonormal matrices preserve the norm and the angle between data vectors. For example, all the results reported in this paper have been computed using this encryption (obfuscation) method. 
Moreover, one can also derive an encrypted (obfuscated) public service by decomposing the secret matrix $Q$ as products of two orthonormal random matrices:
\begin{equation}
Q = S_1 P_1 = S_2 P_2 = ... = S_n P_n = ...
\end{equation}
The matrices $S_n$ and $P_n$ are also $M \times M$ random orthonormal matrices. The orthonormal random matrices $S_n$ can be generated using the same QR approach, however for each 
matrix $S_n$ one should use a different secret seed $s_n$ for the random number generator. 
The $P_n$ matrices are then computed using $P_n = S_n^TQ$, and they are also random and orthonormal, since they are the product of two random orthonormal matrices. 

The host (owner) of the encrypted model can give the $P_n$ matrix to the user $n$, and keep the secret seed $s_n$ such that it can always re-generate the secret matrix $S_n$. 
Then the user $n$ can encrypt the test sample $\tilde{x}$ using $P_n \tilde{x}$ and send the result to the host, which also computes $S_n P_n \tilde{x} = Q \tilde{x} = \tilde{x}'$ before it queries the model. 
In the next step, the host performs the classification of the encrypted $\tilde{x}'$ using the RRNN/RRKN model in the cloud, and returns the class value to the user. This way, the test sample is also encrypted during the transit, and 
the neural network model is rendered useless without knowing the secret matrices $S_n$ and $P_n$. 

\section{Conclusion}

The advantage of the random-SLFNN is that the resulting training task can be formulated as a ridge regression problem. 
Despite their simplicity, many previous studies have indicated that such randomized neural network models can reach a sound classification performance, if they employ a large number of hidden neurons 
relative to the dimensionality of the samples. 
Here we have shown that contrary to this assumption, one can obtain good results even if the number of hidden neurons has the same order of magnitude as the dimensionality of the data samples, 
if the dimensionality of the samples is reasonably high. We have also exploited this idea by developing an efficient iterative training method, 
which significantly improves the classification accuracy. This training approach results in an error correcting neural network architecture, which we called Residual Random Neural Network (RRNN). 
The numerical results are in good agreement with the theoretical results, showing that by increasing the dimensionality of the data space, the random projections drawn from the normal distribution are almost orthogonal.  
This surprising property of high dimensional spaces implicitly results in a much better separation of the randomly projected data samples, and consequently in an improved classification accuracy. 
Following these results we have also extended the residual training algorithm to the least-squares Residual Random Kernel Network (RRKN), achieving a similar classification accuracy performance, with a more simplified architecture. 
Finally, by using orthonormal random projection matrices, we have developed an encryption (obfuscation) method which can be used to protect both the data and the resulted network model. 
The code for the RRNN/RRKN models is provided at: \url{https://github.com/mandrecut/rrnn/}.

\end{document}